\documentclass[conference]{IEEEtran}
\IEEEoverridecommandlockouts
\usepackage{cite}
\usepackage{amsmath,amssymb,amsfonts}
\usepackage{algorithmic}
\usepackage{textcomp}
\usepackage{capt-of}
\usepackage{graphicx}
\usepackage{booktabs}
\usepackage{multirow}
\usepackage{array}
\usepackage{xcolor}
\usepackage{booktabs}
\usepackage{xurl}
\newcommand{\etal}{\emph{et al.}}
\usepackage{bm}

\newlength{\colw}

\newcolumntype{Y}{>{\centering\arraybackslash}p{\colw}}

\usepackage{booktabs}
\usepackage{multirow}
\usepackage{xcolor}

\newcolumntype{Y}{>{\centering\arraybackslash}p{\colw}}

\providecommand{\BibTeX}{{\rm B\kern-.05em{\sc i\kern-.025em b}\kern-.08em T\kern-.1667em\lower.7ex\hbox{E}\kern-.125emX}}
\begin{document}

\title{Learning to Explore Hidden Kinematics for Articulated Object Manipulation


\author{Ruiyao Liu$^{*}$, Boshu Lei$^{*}$, Zhuoyang Pan, Kostas Daniilidis%
\thanks{$^{*}$Equal contribution.}%
\thanks{All authors are with the GRASP Lab, University of Pennsylvania,
Philadelphia, PA 19104, USA.}%
}}

\maketitle

\begin{abstract}
The kinematics of an articulated object is often ambiguous from vision alone.
Interaction resolves the ambiguity, and active perception methods exploit this by searching for the single action that most sharpens a belief over the kinematic parameters at each step.
Such greedy search cannot be extended over a horizon without forward models of the contact and inertial dynamics, which are themselves unknown.
We instead amortize action selection into training.
We maintain a belief distribution over joint type and parameters, initialized from a generative prior and updated by Bayesian filtering on the observed part motion.
To condition the policy on this belief, we render it as a per-point articulation flow field, the motion that the current posterior predicts for every point on the object.
Carrying the inductive bias of articulated motion, this representation generalizes better than a latent encoding of the belief or flow tracked from observation.
We train the policy with reinforcement learning, rewarding the entropy that each interaction removes from the posterior, so that informative exploration becomes learned behavior rather than a search at every step.
Our method outperforms previous approaches across door and drawer manipulation on the PartManip benchmark, and reaches $61.7\%$ success on ArticuRiddle, a new dataset of objects whose appearance implies the wrong articulation, against $44.4\%$ for the best previous method. Project Website: \url{https://hiddenkinematics.github.io/}
\end{abstract}

\section{Introduction}

\begin{figure}
    \centering
    \includegraphics[width=1\linewidth]{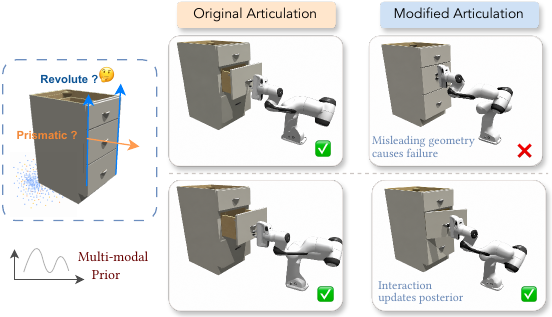}
    \caption{\textbf{Visual appearance can be misleading.} We change a drawer's joint from prismatic to revolute without altering its appearance. Both methods open the original drawer, but on the modified one the baseline is misled by the geometry, whereas ours refines its belief through interaction and adapts.}
    \label{fig:teaser}
\end{figure}

Articulated objects are ubiquitous in everyday environments, whose movable parts are governed by underlying kinematic structures that determine how the object responds to physical interaction.
%
However, geometry alone often leaves the underlying kinematic structure ambiguous.
For example, the cabinet in Fig.~\ref{fig:teaser} admits both a prismatic and a revolute interpretation: in the closed configuration, the two are visually identical, so the distinguishing evidence is absent from the image, and no amount of additional training data or model capacity can recover it. 
The consequence is not merely an estimation error but a manipulation failure: policies trained on large-scale articulated-object datasets commit to a single interpretation and fail on such objects. 
Interaction is what resolves the ambiguity, since the part motion induced by an action carries direct information about the joint that produced it.
The problem is therefore planning under uncertainty: which action to take, given a belief over hypotheses, so that the resulting motion is the most informative about the true structure.

Acting to disambiguate is not a new idea.
Active perception methods maintain a belief over the kinematic structure and search the next interaction to sharpen it, scored by the entropy of a particle filter over candidate models~\cite{hausman2015} or of a hybrid categorical--Gaussian belief~\cite{stefan2014}, or by heuristics such as predicted deformation~\cite{hsaur2023} and affordance~\cite{akm2026}.
Yet all of them commit only to the next action, because scoring an action over a horizon requires a forward model of the contact and inertial dynamics, which a kinematic model alone does not provide.
Model-free reinforcement learning provides a general way to learn long-horizon plans, but the challenge is how to expose the belief to the policy due to the incompatibility of the belief in joint space and the observation in Cartesian space.
Our insight is that every hypothesis predicts a motion for each point on the object, so the belief can instead be rendered as a pointwise articulation flow field that is aligned with the observation. 
Training the policy conditioned on the flow field with an information-seeking reward turns the interaction into a learned behavior rather than a search at every step.

This motivates us to develop a closed-loop framework that allows the robot to learn to explore the articulation with its understanding of the object.
We maintain a belief distribution over plausible kinematic configurations, initialized from a learned kinematic prior and recursively refined according to how well each hypothesis explains the observed part motion.
To condition the policy on the belief, we render the belief into a point-wise articulation flow that encodes the motion predicted by the current kinematic estimate, which provides the policy with a compact spatial representation of accumulated interaction information.
Because the filter maintains this posterior explicitly at every step, the information an interaction yields is the entropy that the interaction removes from the posterior.
We use this quantity as a reward alongside the task reward, so that the policy is trained to induce motion informative about the unknown articulation while making progress toward the manipulation goal.
Together, belief refinement, articulation flow, and informative interaction allow the policy to adapt as new evidence about the object's articulation becomes available.

We evaluate our approach on the PartManip benchmark and introduce the ArticuRiddle dataset, designed to evaluate manipulation under visually ambiguous articulation.
Our method consistently outperforms existing approaches across door and drawer manipulation tasks and achieves $61.7\%$ success on ArticuRiddle, compared with $32.4\%$ for PartManip.
We further demonstrate zero-shot transfer to real-world articulated objects, where the articulation estimate and corresponding flow field are progressively refined through interaction.

Our main contributions are:
\begin{itemize}

    \item We introduce a posterior-derived articulation flow that converts the evolving articulation belief into a spatial representation for the manipulation policy, allowing accumulated interaction information to guide actions.

    \item We compute the entropy reduction of the posterior and use it as a reinforcement learning objective that encourages the robot to generate informative motion about the unknown articulation during manipulation.

    \item We introduce the ArticuRiddle dataset for evaluating manipulation under visually ambiguous articulation, and demonstrate improvements over existing methods in simulation with zero-shot transfer to real-world objects.
\end{itemize}

\section{Related Work}

\emph{Articulation Estimation and Manipulation.} Estimating articulated-object kinematics from point clouds or videos has been widely studied~\cite{jiang2022ditto}. Explicit methods segment rigid parts and identify their joints~\cite{katz2008}, with Bayesian extensions maintaining a posterior over possible kinematic structures~\cite{sturm2011}. Recent methods also predict actionable quantities such as grasp poses~\cite{gamma2024}, or construct physics models for manipulation planning~\cite{sim2real22023}. In parallel, implicit methods learn manipulation policies directly through imitation or reinforcement learning~\cite{geng2023partmanip,wang2025articubot,do2025watchless}. Although avoiding explicit planning, they typically rely on articulation estimates obtained before interaction and are thus vulnerable when visually similar objects have different kinematics. Prior work connects articulation estimates to manipulation using several representations. Affordance and actionability maps predict where and how to interact on 3D points~\cite{sim2real22023}, image pixels~\cite{where2act}, or trajectory proposals~\cite{vatmart2022}. Dense motion fields predict per-point motion as articulation flow~\cite{eisner2022flowbot3d,li2024flowbothd} or interaction-derived scene flow~\cite{nie2023sfa}, while other methods condition policies on compact latents~\cite{do2025watchless}. Unlike fields predicted once from a static observation, our flow is derived from the current posterior and recomputed after each interaction, conveying motion evidence to the policy.

\emph{Active Perception for Articulation.}
A separate line of work treats articulation estimation as an active perception problem, choosing actions that make the resulting motion more informative about the object's structure. 
These approaches update a belief distribution for the kinematic parameters through interactions and plan actions to gather information.
The entropy of the posterior is estimated from the samples of a particle filter~\cite{hausman2015}, or computed analytically over a hybrid belief that is categorical in joint type and Gaussian in the continuous joint parameters~\cite{stefan2014}.
Other approaches use deformation~\cite{hsaur2023}, affordance~\cite{akm2026}, and affordance prediction variance~\cite{wang2022adaafford} as heuristics to select actions.
However, they plan the next action, which is myopic. 
To plan non-myopically, one can use reinforcement learning to optimize expected return from information yielded by the whole trajectory, with action selection amortized into training rather than searched at every step.

\emph{Reinforcement Learning under Partial Observability.}
Articulated-object manipulation is a partially observable Markov decision process (POMDP), with kinematic parameters and deformation as hidden states. 
Online POMDP solvers maximize expected belief-state returns through tree search~\cite{ross2008online} and Monte Carlo methods~\cite{silver2010mc,zachary2018}. 
Information seeking can instead be incorporated as expected belief information gain~\cite{rein2016,mauricio2010}. 
In practice, this term is approximated: ASID~\cite{asid2024} rewards Fisher information about dynamics parameters, Curtis~\etal~\cite{curtis2023} reward observations that separate particle-filter hypotheses, and Xie~\etal~\cite{xie2023} use 2D segmentation uncertainty as a surrogate. For articulation manipulation, VAT-MART~\cite{vatmart2022} encourages its state policy to explore trajectories scored poorly by the perception model, but uses this policy only during training. 
Unlike these approximations, our particle filter maintains an explicit articulation posterior at every step, allowing us to directly use its entropy reduction after each interaction as the information-gain reward.

\begin{figure*}[!t]
    \centering
    \includegraphics[width=1\linewidth]{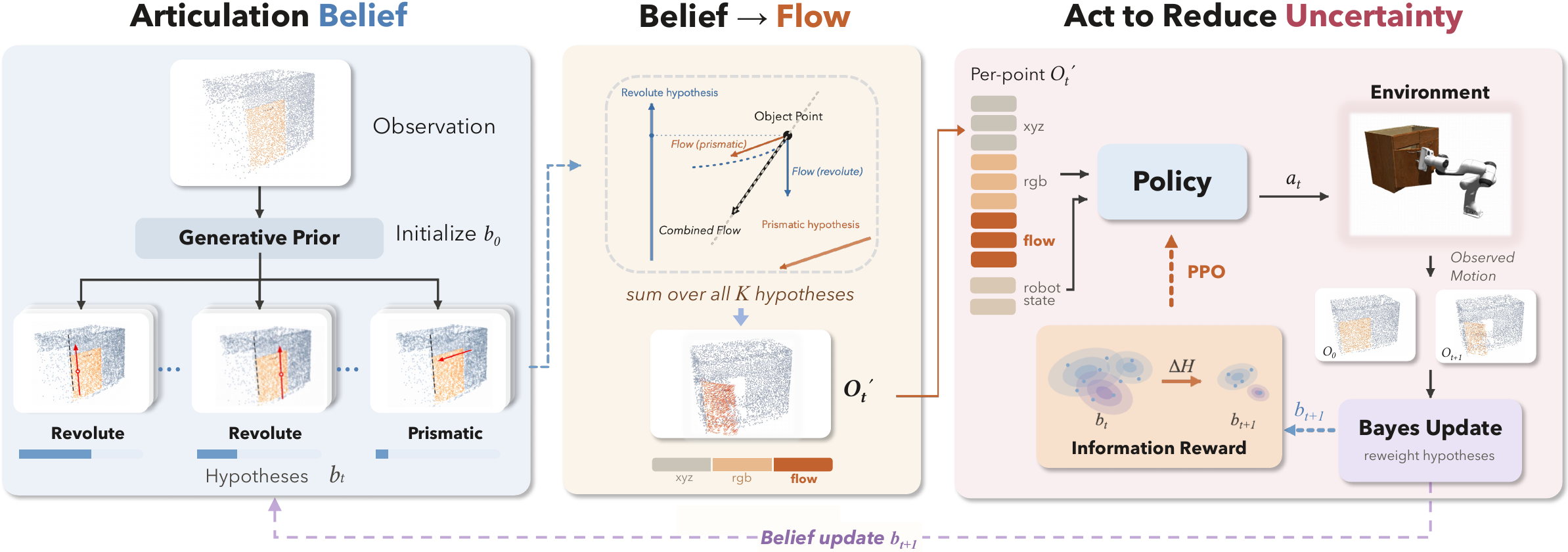}
      \vspace{-1em}
    \caption{\textbf{Pipeline Overview.} Given the initial observation, we use a generative prior to initialize a set of particles, where each particle represents a hypothesis of the object’s articulation model. At each interaction step, each hypothesis is converted into a per-point flow field. These flow fields are weighted-averaged over the particles. The flow field is first concatenated with the current point cloud as $O_t'$ and then fused with the encoded robot’s proprioceptive state as input to the actor. The predicted action is executed in the environment, and the observed motion is used to update the particle weights through a Bayes filter. The information reward, defined as the reduction in belief entropy from before to after the update, is combined with the task reward to train the policy using PPO.}
    \label{fig:pipeline}
\end{figure*}

\section{Method}
We consider the task of manipulating an articulated object with an unknown kinematic structure while estimating its articulation online from interaction. 
The agent acts on a target part and uses the resulting motion to infer its joint type and parameters. 
The overall pipeline is shown in Fig.~\ref{fig:pipeline}.

\subsection{Online Articulation Estimation}
\label{sec:articulation}
\subsubsection{Generating Hypothetical Articulations}
\label{sec:hypothesis}
Given the observed point cloud $O_t$ of a target object with part
segmentation, we separate the object into part-centric point clouds
$O_t^1,O_t^2,\ldots,O_t^N$, where $N$ is the number of object parts.
We focus on a target part $O_t^i$ and infer its underlying
articulation throughout interaction. To initialize the belief distribution, we sample articulation
hypotheses from NAP~\cite{DBLP:journals/corr/abs-2305-16315}, a
generative prior conditioned on the initial observation that yields
candidate joint parameters. The $k$-th candidate, and hence particle,
is denoted by $\theta^{(k)}=(c^{(k)},d^{(k)})$, where
$c^{(k)}\in\{\mathrm{r},\mathrm{p},\mathrm{f}\}$ denotes a revolute,
prismatic, or fixed joint. Its parameters are
$d^{(k)}=(u^{(k)},p^{(k)})$, $u^{(k)}$, and $\varnothing$ for
revolute, prismatic, and fixed joints, respectively, where $u^{(k)}$
and $p^{(k)}$ denote the axis direction and pivot. The initial
particle pool $\{\theta^{(k)}\}_{k=1}^{K}$, with
$\theta^{(k)}\sim p(\theta\mid O_0)$, captures the uncertainty over
how the target part may articulate. The
initial weights $w_k$ are obtained by normalizing the NAP
confidence scores.

\subsubsection{Updating Hypotheses from Interaction}
\label{sec:update}
After applying action $a_t$, we obtain the resulting motion observation
$\mathcal{D}_t=\{(\mathbf{x}_j,\mathbf{x}'_j)\}_{j=1}^{M}$, where
$\mathbf{x}_j$ and $\mathbf{x}'_j$ denote the position of the $j$-th
tracked point on the target part before and after interaction,
respectively, for $M$ tracked points, and evaluate how well it can be
explained by each articulation hypothesis. A revolute hypothesis fits
the observed motion as rotation about its candidate axis, whereas a
prismatic hypothesis fits it as translation along the candidate axis.
A fixed hypothesis instead expects negligible object motion.

Let $q_{i,t}$ denote the joint coordinate of the target part at time
$t$, and $\delta q_{i,t}=q_{i,t+1}-q_{i,t}$ the displacement it
undergoes under action $a_t$; we estimate $\delta q_{i,t}$ jointly
with $\theta^{(k)}$ from the same correspondences $\mathcal{D}_t$ using minimum square fitting.
Assuming isotropic Gaussian observation noise, we define the
likelihood weight of particle $\theta^{(k)}$ as
\begin{equation}
\begin{aligned}
w_k
&=
p(\mathcal{D}_t \mid \theta^{(k)}, \delta q_{i,t}) \\
&\propto
\exp\left(
-\sum_{j=1}^{M}
\left\|
(\mathbf{x}'_j-\mathbf{x}_j)
-
\mathbf{f}_j(\theta^{(k)},\delta q_{i,t})
\right\|_2^2
/2\sigma^2
\right).
\end{aligned}
\label{eq:likelihood}
\end{equation}
where $\mathbf{f}_j(\theta^{(k)},\delta q_{i,t})$ denotes the
displacement of point $j$ prescribed by the kinematic constraint of
hypothesis $\theta^{(k)}$, and $\sigma^2$ is the observation noise
variance. Hypotheses that better fit the observed motion therefore
receive larger likelihood weights.

The articulation belief is updated recursively as
\begin{equation}
\begin{aligned}
    p(\theta^{(k)}\mid O_{0:t+1},a_{0:t})
    &\propto
    p(\mathcal{D}_t\mid \theta^{(k)},\delta q_{i,t})\,
    p(\theta^{(k)}\mid O_{0:t},a_{0:t-1}) \\
    &=
    w_k\,
    p(\theta^{(k)}\mid O_{0:t},a_{0:t-1}) .
\end{aligned}
\label{eq:posterior}
\end{equation}
After normalization, the posterior weights $w_k$ are used as the
resampling probabilities. The second factor represents the belief
accumulated through previous interactions. We implement this recursive
inference using particle filtering with importance weighting and
resampling. 

\subsection{Flow Fields from Belief Distribution}
\label{sec:flow}
\begin{figure}
    \centering
    \includegraphics[width=1\linewidth]{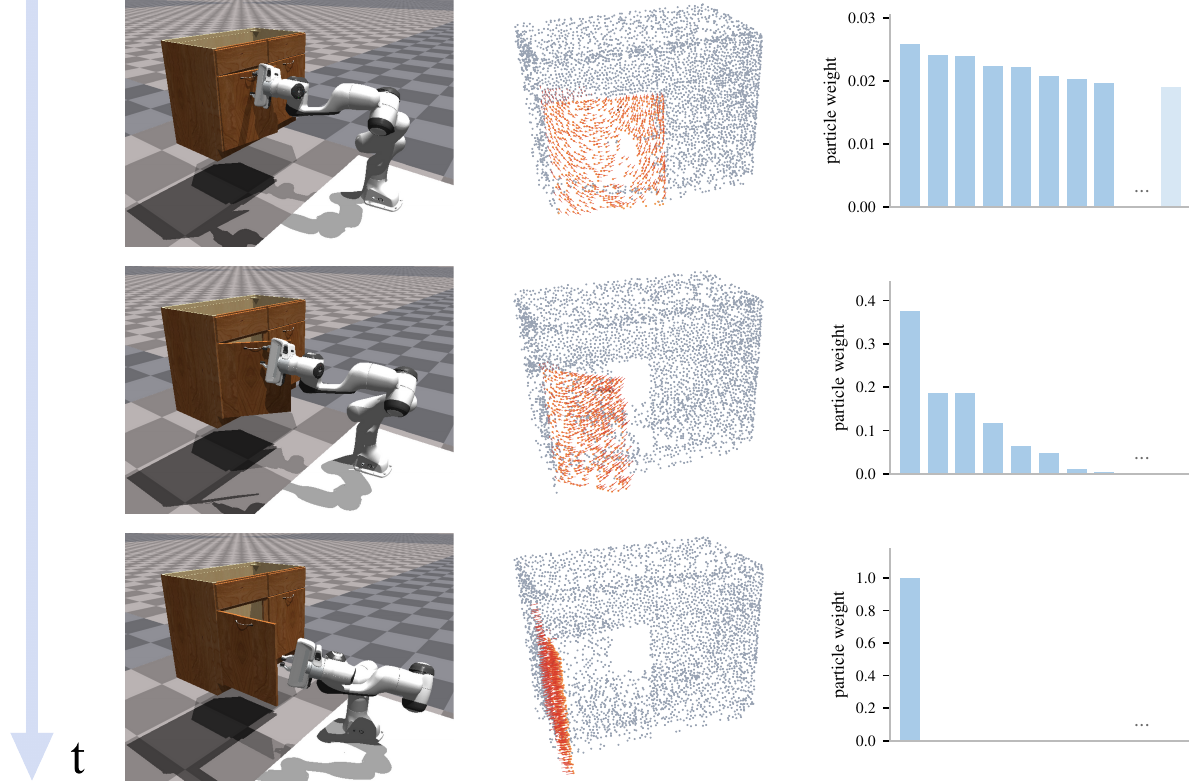}
    \caption{\textbf{Flow field updates during interaction at time $t$.} As the robot
interacts with the object, hypotheses inconsistent with the observed motion are
suppressed and the posterior gradually concentrates. The flow is updated
accordingly, providing the policy with increasingly accurate motion cues.}
    \label{fig:flow}
\end{figure}
As the agent interacts with an articulated object, its unknown
kinematic structure is gradually revealed through the resulting
motion. Planning directly from the current articulation estimate can
make the trajectory sensitive to estimation errors and limit adaptive
correction, while relying solely on interaction history requires the
policy to implicitly infer and retain the evolving articulation
structure. We summarize the interaction into a compact
representation that carries the current articulation estimate into
the policy observation directly.

As shown in Fig.~\ref{fig:flow}, at each time step, we convert the current posterior belief $\{(\theta^{(k)},w_k)\}_{k=1}^{K}$ over the target part into a per-point motion field according to the kinematic constraints represented by the particles. 
Under a revolute hypothesis, the motion is tangent to the circle traced by the point about the axis, while under a prismatic hypothesis, it points along the axis direction. 
We normalize the motion field of each joint type by its mean magnitude over the part, so a revolute field retains a mild gradient that grows with distance from the axis, while a prismatic field remains uniform. 
The normalized fields are aggregated pointwise by a vector sum according to the posterior belief. 
We append this displacement to every point's feature, augmenting it from $\mathbb{R}^{6}$ to $\mathbb{R}^{9}$, and denote the resulting point cloud by $O_t'$, which is observed by the policy. 
Because this observation is recomputed from the belief at every step, it serves as an implicit memory of past interactions, carrying accumulated information into the current observation without requiring the policy to retain interaction history.
\subsection{Reinforcement Learning for Informative Interaction}
\label{sec:rl}

We use reinforcement learning as the optimizer of our system. Since the particle filter of Sec.~\ref{sec:update} maintains $p(\theta\mid O_{0:t},a_{0:t-1})$ at every step, the information gain about $\theta$ requires no proxy on the observed motion: it is the entropy that an interaction removes from that posterior.

\paragraph{Entropy of the articulation belief}
We write the belief at time $t$ as the weighted particle set
$b_t=\{(\theta^{(k)},w_k)\}_{k=1}^{K}$ with
$\theta^{(k)}=(c^{(k)},d^{(k)})$. 
We keep the type factor exact and estimate the parameter term separately inside each type branch,
\begin{equation}
    \mathbf{H}[b_t]
    =
    \mathbf{H}\!\left[P_t(c)\right]
    +
    \sum_{c\in\{r,p\}} P_t(c)\,
    \widehat{\mathbf{H}}\!\left[b_t(d\mid c)\right],
    \label{eq:belief_entropy}
\end{equation}
where $P_t(c)=\sum_{k:\,c^{(k)}=c} w_k$ is the posterior mass on type $c$. 
A prismatic particle carries an axis direction
$d=\mathbf{u}\in\mathbb{S}^{2}$ and a revolute one direction and pivot $d=(\mathbf{u},\mathbf{p})\in\mathbb{R}^{6}$; a fixed one has no parameters and contributes only the type term.

\paragraph{Kernel density estimate of the branch entropy}
The particle weights alone are not sufficient. When the particles have
concentrated around the true axis their weights become similar, which
makes the weight entropy $-\sum_k w_k\log w_k$ large exactly when the
belief has converged; it reflects how many hypotheses survive, not how
far apart they lie. Following Hausman~\etal~\cite{hausman2015}, we
instead estimate the differential entropy of each branch
non-parametrically, from the positions of the particles. Renormalizing
inside the branch, $\bar w_k = w_k/P_t(c)$, we place a Gaussian kernel
on every particle,
\begin{equation}
\begin{gathered}
\hat f_{\mathbf{B}}(d)
= \sum_{k:\,c^{(k)}=c} \bar w_k\,
K_{\mathbf{B}}\!\left(d-d^{(k)}\right), \\[2pt]
K_{\mathbf{B}}(x)
= |\mathbf{B}|^{-1/2}K\!\left(\mathbf{B}^{-1/2}x\right),
\quad
K(x)
= (2\pi)^{-D/2}e^{-\frac{1}{2}x^{\top}x},
\end{gathered}
\label{eq:kde}
\end{equation}
where $D$ is the dimension of $d$ in the branch ($3$ for prismatic, $6$
for revolute) and the bandwidth follows Scott's rule~\cite{scott1979optimal},
\begin{equation}
    \mathbf{B}_{ij}=0 \;\; (i\neq j),
    \qquad
    \sqrt{\mathbf{B}_{ii}} = n_c^{-\frac{1}{m_c+4}}\,\sigma_i ,
    \label{eq:scott}
\end{equation}
with $\sigma_i$ the weighted standard deviation of the $i$-th coordinate
within the branch, $n_c$ the number of particles it holds, and $m_c$ the
intrinsic dimension of the constrained parameters. Evaluating the
estimate at the particle locations themselves gives the resubstitution
estimate~\cite{ahmad1976nonparametric} of the branch entropy,
\begin{equation}
    \widehat{\mathbf{H}}\!\left[b_t(d\mid c)\right]
    =
    -\sum_{k:\,c^{(k)}=c} \bar w_k \, \log \hat f_{\mathbf{B}}\!\left(d^{(k)}\right).
    \label{eq:resub}
\end{equation}
Both the mixture and the outer average are weighted by $\bar w_k$, where
\cite{hausman2015} uses $1/n_c$; the weighted form reduces to theirs
when the weights within a branch are uniform. The branch entropies also
enter Eq.~\eqref{eq:belief_entropy} with the type term, so resolving the
joint type counts as a reduction in uncertainty.

\paragraph{Information gain as a reward}
The information gain collected at step $t$ is the drop in
Eq.~\eqref{eq:belief_entropy} produced by that step's observation,
\begin{equation}
R_{\mathrm{info},t}
=
\mathbf H[b_t]
-
\mathbf H[b_{t+1}].
     \label{eq:info_reward}
\end{equation}
It is a difference of entropies rather than a negated entropy, and this
matters. A reward proportional to $-\mathbf{H}[b_{t+1}]$, or to per-step
measure of surprise, remains available after the belief has converged,
so the policy can keep collecting it while standing still. The
difference in Eq.~\eqref{eq:info_reward} vanishes once the belief stops
moving, so each nat of information can be earned only once.

\paragraph{PPO Reward}
We optimise the policy with PPO~\cite{schulman2017proximal} using a reward that combines the task and information terms:
\begin{equation}
    R = R_{\mathrm{task}} + \lambda_i\, R_{\mathrm{info}},
    \label{eq:reward}
\end{equation}
where $\lambda$ balances the two terms. $R_{\mathrm{task}}$ is the
part-aware manipulation reward,
\begin{equation}
    R_{\mathrm{task}}
    =
    \lambda_r R_{\mathrm{rot}} + \lambda_d R_{\mathrm{dist}}
    + \lambda_s R_{\mathrm{succ}} ,
    \label{eq:task_reward}
\end{equation}
$R_{\mathrm{rot}}$ rewards aligning the gripper axis with the target
part and $R_{\mathrm{dist}}$ penalizes the gripper's distance to the
handle centre, while $R_{\mathrm{succ}}$ is paid once when the part is
sufficiently open. 

%

Training the policy from scratch is expensive; we thus train a state-based expert $\pi_{\mathrm{expert}}$ with privileged access to the object state and distill it into a vision-based policy on the states the student itself visits, following~\cite{geng2023partmanip},
\begin{equation}
    \mathcal{L}_{DA}
    =
    \frac{1}{\lvert D_{\pi_\phi} \rvert}
    \sum_{o,\, s \, \in \, D_{\pi_\phi}}
    \left\lVert
      \pi_{\mathrm{expert}}(s) - \pi_{\mathrm{student}}(o)
    \right\rVert_2 .
    \label{eq:dagger}
\end{equation}
We continue to train the student with reinforcement learning after distillation, under the observations it will be deployed with, retaining the expert as a regularizer:
\begin{equation}
    \max_{\phi} \; \mathcal{L}_{\mathrm{ppo}} \; - \; \lambda_{\mathrm{DA}}\, \mathcal{L}_{DA} .
    \label{eq:objective}
\end{equation}
%
\section{Experiments}
\label{sec:exp}

\subsection{Simulation Settings}
\label{sec:exp-setup}

We simulate in IsaacGym~\cite{makoviychuk2021isaac} with a Franka Panda arm and parallel gripper. The observation comprises a $4000$-point colored point cloud of the workspace, fused from three depth cameras, in which the target part is indicated by its center, and a $41$-dimensional proprioceptive vector containing the joint states, the robot base  position, end-effector pose, and target-part center. The policy outputs a target end-effector pose and two finger targets, which are converted into joint targets by damped least-squares IK.

\begin{figure}
\centering
\includegraphics[width=1\linewidth]{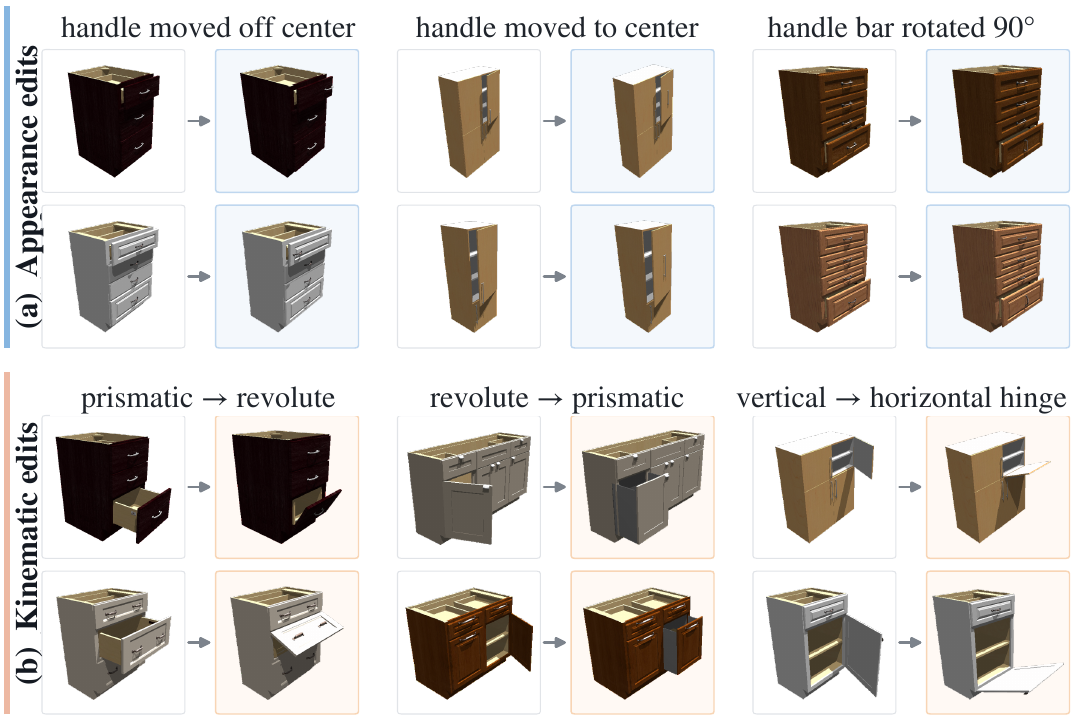}
\caption{\textbf{Overview of ArticuRiddle.} We construct challenging objects through (a) handle-pose edits and (b) joint-parameter edits that preserve appearance, both testing manipulation under ambiguous geometric cues.}
\label{fig:dataset}
\vspace{-4mm}
\end{figure}

We evaluate on two benchmarks. \textbf{PartManip}~\cite{geng2023partmanip} contains articulated objects from GAPartNet ~\cite{DBLP:conf/cvpr/GengXZ0YHW23}: $363/63$ training/validation doors and $200/40$ drawers. Our \textbf{ArticuRiddle} contains $100$ objects derived from PartManip (Fig.~\ref{fig:dataset}). Half have altered handle positions or orientations; the others have modified joint axes but unchanged appearance, forming visually similar pairs with different kinematics.

All learning-based methods are jointly trained on doors and drawers. For each object, we draw $K=200$ articulation hypotheses from NAP to
initialize the particle belief. The state expert uses the PartManip reward~\cite{geng2023partmanip}; the vision policy uses Eq.~\eqref{eq:reward}\eqref{eq:task_reward} with $\lambda_r=0.2$, $\lambda_d=2$, $\lambda_s=100$, and $\lambda_i=0.1$. $R_{\mathrm{succ}}$ is awarded once the joint coordinate exceeds $0.7$rad for doors or $0.7$m for drawers. The PPO actor/critic learning rates are $10^{-4}/3\times10^{-4}$, and the DAgger learning rate is $3\times10^{-4}$.

We report mean and standard deviation over three independently trained seeds for learning-based methods and three complete evaluations for planning-based methods. Success requires opening a door beyond $30^\circ$ or a drawer beyond $0.2$,m. Articulation estimation uses joint-type accuracy and $\mathrm{ASR}_x$: revolute predictions require axis-angle and pivot errors below $x^\circ$ and $x$cm, respectively, whereas prismatic predictions require only the axis-angle criterion. We report $x\in\{10,20\}$.

We compare against four prior methods spanning explicit
estimate-then-plan systems and learned manipulation policies.
\begin{table*}[t]
\centering
\caption{\textbf{Articulation estimation}on PartManip and ArticuRiddle.}
\label{tab:articulation_estimation}

\setlength{\tabcolsep}{4pt}
\renewcommand{\arraystretch}{1.2}
\footnotesize

\begin{tabular*}{\textwidth}{@{\extracolsep{\fill}}llccccccccc@{}}
\toprule

\multirow{2}{*}{\textbf{Category}}
& \multirow{2}{*}{\textbf{Method}}
& \multicolumn{3}{c}{\textbf{Train}}
& \multicolumn{3}{c}{\textbf{Val}}
& \multicolumn{3}{c}{\textbf{ArticuRiddle}} \\

\cmidrule(lr){3-5}
\cmidrule(lr){6-8}
\cmidrule(lr){9-11}

& &
Acc.$\uparrow$ &
ASR$_{10}\uparrow$ &
ASR$_{20}\uparrow$ &
Acc.$\uparrow$ &
ASR$_{10}\uparrow$ &
ASR$_{20}\uparrow$ &
Acc.$\uparrow$ &
ASR$_{10}\uparrow$ &
ASR$_{20}\uparrow$ \\

\midrule

\multirow{3}{*}{Door}

& AKM~\cite{akm2026}
& 79.0$\pm$1.9 & 48.2$\pm$1.3 & 55.9$\pm$2.5
& 84.1$\pm$1.6 & 29.6$\pm$5.6 & 33.3$\pm$5.9
& 64.0$\pm$3.5 & 30.7$\pm$6.1 & 32.7$\pm$5.8 \\

& H-SAUR~\cite{hsaur2023}
& 81.7$\pm$1.9 & 40.3$\pm$3.6 & 46.6$\pm$4.6
& 85.2$\pm$2.7 & 35.1$\pm$5.4 & 43.2$\pm$8.1
& 68.1$\pm$1.9 & 39.0$\pm$2.9 & 50.3$\pm$2.9  \\

& Ours
& \textbf{82.7}$\boldsymbol{\pm}$\textbf{1.3}
& \textbf{62.4}$\boldsymbol{\pm}$\textbf{1.3}
& \textbf{72.7}$\boldsymbol{\pm}$\textbf{2.1}
& \textbf{87.8}$\boldsymbol{\pm}$\textbf{2.2}
& \textbf{59.2}$\boldsymbol{\pm}$\textbf{5.1}
& \textbf{70.1}$\boldsymbol{\pm}$\textbf{3.0}
& \textbf{78.4}$\boldsymbol{\pm}$\textbf{5.2}
& \textbf{57.7}$\boldsymbol{\pm}$\textbf{5.6}
& \textbf{65.4}$\boldsymbol{\pm}$\textbf{3.3} \\

\midrule

\multirow{3}{*}{Drawer}


& AKM~\cite{akm2026}
& 75.9$\pm$1.6 & 42.1$\pm$1.9 & 49.7$\pm$3.8
& 79.2$\pm$4.4 & 41.5$\pm$4.0 & 52.5$\pm$6.5
& 54.1$\pm$4.7 & 35.8$\pm$2.2 & 41.7$\pm$1.8 \\

& H-SAUR~\cite{hsaur2023}
& 73.0$\pm$2.0 & 43.0$\pm$2.3 & 49.7$\pm$0.4
& 73.5$\pm$1.9 & 38.1$\pm$0.3 & 54.3$\pm$7.1
& 59.0$\pm$1.3 & 29.0$\pm$5.5 & 36.2$\pm$3.3 \\

& Ours
& \textbf{89.6}$\boldsymbol{\pm}$\textbf{1.2}
& \textbf{61.4}$\boldsymbol{\pm}$\textbf{2.0}
& \textbf{69.2}$\boldsymbol{\pm}$\textbf{2.3}
& \textbf{89.0}$\boldsymbol{\pm}$\textbf{1.0}
& \textbf{43.5}$\boldsymbol{\pm}$\textbf{4.8}
& \textbf{58.3}$\boldsymbol{\pm}$\textbf{4.3}
& \textbf{94.4}$\boldsymbol{\pm}$\textbf{3.2}
& \textbf{55.8}$\boldsymbol{\pm}$\textbf{2.5}
& \textbf{61.8}$\boldsymbol{\pm}$\textbf{1.4} \\

\bottomrule
\end{tabular*}
\end{table*}
\begin{itemize}
    \item \textbf{H-SAUR}~\cite{hsaur2023} maintains a particle-based belief over articulation models. It reconstructs objects in a physics simulator, evaluates single-point contact under these hypotheses, and executes an interaction producing the largest displacement. The outcome is then used to update the hypothesis weights before the next action is selected.
    
    \item \textbf{AKM}~\cite{akm2026} predicts the affordance map using DIFT~\cite{DIFT} feature cosine similarity from a reference image retrieved through RAM~\cite{RAM}. It iteratively selects the contact point on the 2D image with the highest score and updates the affordance map after each interaction. 
    
    \item \textbf{Wang~\etal~}\cite{wang2024vtae} tracks the target part across RGB-D observations and fits its joint axis from the observations. It replans the manipulation trajectory online whenever the estimate is updated. It represents articulation with a single estimate rather than a distribution over plausible configurations.

    \item \textbf{PartManip}~\cite{geng2023partmanip} learns a cross-category manipulation policy directly from point-cloud observations. A privileged state policy is first trained as an expert and then distilled into a visual student policy through Dagger. 
\end{itemize}
\begin{table}[!t]
\centering
\caption{\textbf{Manipulation success rate.} 
}
\label{tab:manipulation_success}
\setlength{\tabcolsep}{2.5pt}
\renewcommand{\arraystretch}{1.2}
\scriptsize

\begin{tabular*}{\columnwidth}{@{\extracolsep{\fill}}lccccc@{}}
\toprule
\multirow{2}{*}{\textbf{Method}}
& \multicolumn{2}{c}{\textbf{Door}}
& \multicolumn{2}{c}{\textbf{Drawer}}
& \multirow{2}{*}{\textbf{ArticuRiddle}} \\
\cmidrule(lr){2-3}\cmidrule(lr){4-5}
& Train & Val
& Train & Val & \\
\midrule

H-SAUR~\cite{hsaur2023}
& 21.8$\pm$1.2
& 27.0$\pm$1.3
& 33.5$\pm$1.8
& 32.7$\pm$1.5
& 44.4$\pm$2.2 \\

AKM~\cite{akm2026}
& 25.3$\pm$2.0
& 19.0$\pm$1.6
& 26.7$\pm$1.1
& 22.6$\pm$1.4
& 27.0$\pm$2.5 \\

Wang~\etal~\cite{wang2024vtae}
& 37.4$\pm$2.5
& 39.7$\pm$5.4
& 44.2$\pm$3.3
& 27.8$\pm$10.6
& 18.2$\pm$3.9 \\

PartManip~\cite{geng2023partmanip}
& 44.5$\pm$0.7
& 40.6$\pm$2.7
& 67.1$\pm$2.6
& 74.3$\pm$4.3
& 32.4$\pm$1.4 \\

\midrule

Obs-flow
& 59.4$\pm$1.9
& 43.6$\pm$3.6
& 76.7$\pm$1.8
& 88.0$\pm$1.1
& 53.8$\pm$3.4  \\

Latent
& 60.5$\pm$1.1
& 41.2$\pm$9.0
& 66.4$\pm$4.2
& 87.7$\pm$3.5
& 36.2$\pm$1.1 \\

NAP w/o post.
& 62.1$\pm$1.9
& 41.4$\pm$1.1
& 80.3$\pm$2.4
& 88.3$\pm$1.6
& 48.0$\pm$2.6 \\

Ours
& \textbf{68.2}$\boldsymbol{\pm}$\textbf{1.3}
& \textbf{51.3}$\boldsymbol{\pm}$\textbf{1.9}
& \textbf{89.0}$\boldsymbol{\pm}$\textbf{3.0}
& \textbf{95.4}$\boldsymbol{\pm}$\textbf{1.5}
& \textbf{61.7}$\boldsymbol{\pm}$\textbf{1.1} \\

\bottomrule
\end{tabular*}
\end{table}

We implement H-SAUR and AKM using the same objects, target-part masks, and evaluation criteria, and jointly retrain PartManip on doors and drawers using our split and budget. Table~\ref{tab:manipulation_success} shows that ours performs best throughout, improving over Wang~\etal~\cite{wang2024vtae} from $37.4\%$ to $68.2\%$ on training doors and from $44.2\%$ to $89.0\%$ on training drawers. It also outperforms PartManip by $23.7/10.7$ points on door train/validation and $21.9/21.1$ on drawer train/validation. Their shared base policy and distillation procedure isolate the benefit of inferred kinematics. The largest gap occurs on ArticuRiddle ($61.7\%$ vs.\ $32.4\%$), demonstrating the value of interaction when appearance suggests incorrect articulation. Table~\ref{tab:articulation_estimation} further shows that our method consistently outperforms AKM and H-SAUR in joint-type accuracy and $\mathrm{ASR}_{10/20}$ across all splits.

\subsection{Articulation Representation}
\label{sec:exp-repr}

Because the belief contains abstract kinematic parameters while the policy acts on point clouds, its representation determines how effectively it guides manipulation. We compare three alternatives under the same policy architecture, reward, and training budget.

\begin{itemize}
\item \textbf{Obs-flow.} We append raw per-point displacement between consecutive observations. It matches our field's dimensionality but represents only the latest action, is zero before contact or without motion, and is dominated by tracking noise for small displacements.

\item \textbf{NAP w/o post.} We sample particles from NAP prior~\cite{DBLP:journals/corr/abs-2305-16315} and freeze their hypotheses and weights, deriving the flow from the same estimate throughout the episode.

\item \textbf{Latent.} We replace the flow channels with a compact latent. A privileged MLP maps a $9$-dimensional input (pivot ($3$), axis ($3$), joint position and velocity ($2$), and binary door/drawer type ($1$))to a $16$-dimensional latent $z$, followed by non-affine LayerNorm to fix $|z|$ and prevent collapse. A temporal 1-D CNN predicts $\tilde z$ from a $10$-step history of proprioception, previous actions, and the same noisy articulation estimate supplied by our particle filter. Following~\cite{do2025watchless}, the encoder, adaptation module, and policy are jointly trained with our recipe and a latent-matching regularizer between $z$ and $\tilde z$; the policy receives $z$ during training and only $\tilde z$ at evaluation.
\end{itemize}

Table~\ref{tab:manipulation_success} separates these components. Relative to NAP w/o post., posterior updating improves door train, door validation, and ArticuRiddle success by $6.1$, $9.9$, and $13.7$ points, respectively. Obs-flow is weaker because it is unavailable before motion and unreliable for small single-step displacements. The latent variant also generalizes poorly to ArticuRiddle ($36.2\%$): it must learn from training objects how the axis constrains each contact point, whereas our flow computes this relation explicitly for any geometry.

\subsection{Ablation Studies}
\label{sec:exp-ablation}

We ablate the interaction signal while keeping all other settings fixed. The first variant uses joint displacement $\delta q$, the opening term provided by the task reward, with the coefficient from PartManip~\cite{geng2023partmanip}. The other two replace this term with either the Kullback--Leibler divergence $D_{\mathrm{KL}}$ between the beliefs before and after each update or our entropy decrease $\Delta H$ in Eq.~\eqref{eq:info_reward}. Both information rewards are standardized before weighting. For $\Delta H$ and $D_{\mathrm{KL}}$, we sweep
$\lambda_i \in \{0.05,0.1,0.2\}$ and report the best setting for each. We also evaluate \emph{w/o RL}, trained solely through distillation.

Table~\ref{tab:info_reward_ablation} shows that $\Delta H$ performs best in every column. This differs from~\cite{hausman2015}, where these criteria select only the next action; here, they reward each interaction outcome and are accumulated by the policy over the episode. The belief depends not only on the articulation but also on the joint state, observation process (e.g., camera viewpoint and point tracking), and particle filter. Its entropy therefore reflects the agent's overall reasoning during interaction.

\begin{figure}
    \centering
    \includegraphics[width=1\linewidth]{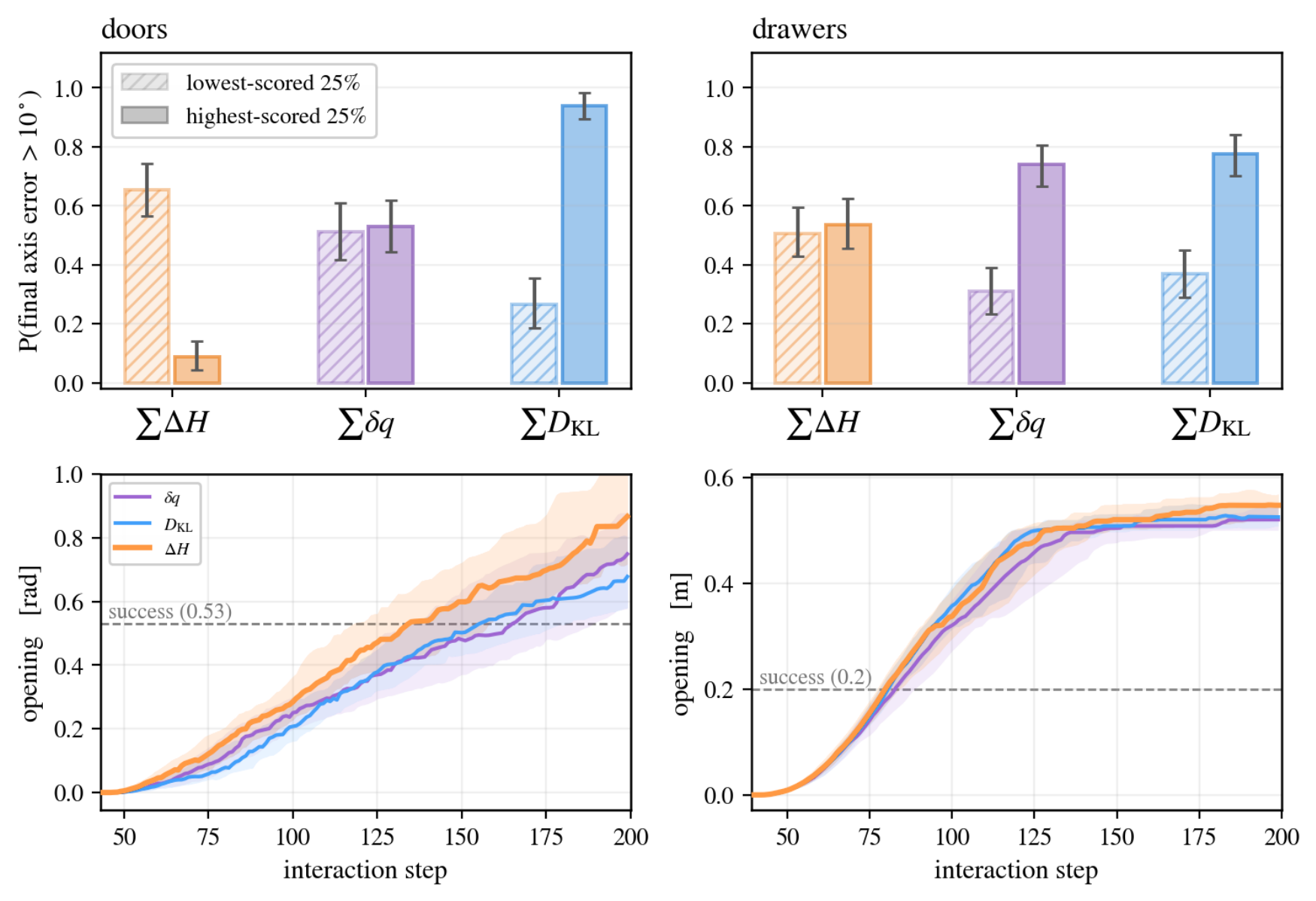}
    \caption{\textbf{Reward alignment and opening progress on the training set.}
  \textbf{Top}: episodes collected with our trained policies are scored by three different rewards; bars give the
  fraction with a final axis error above $10^{\circ}$ in the lowest- (hatched) and
  highest-scoring (solid) quarter. The results suggest optimizing the entropy-reduction return may lead to more accurate estimates.
  \textbf{Bottom}: median opening over assets for the policy trained with each
  reward; the dashed line marks the success threshold.
  Error bars and bands are bootstrapped $95\%$ confidence intervals, over episodes
  (top) and over assets (bottom).}
    \label{fig:reward_alignment}
    \vspace{-4mm}
\end{figure}



Fig.~\ref{fig:reward_alignment} (top) evaluates each reward against final axis error, which none directly uses. Ranking episodes by accumulated reward, $\sum_t\Delta H$ reduces the fraction of doors ending above $10^\circ$ from $0.65$ in the lowest-scoring quarter to $0.09$ in the highest, whereas $\sum_tD_{\mathrm{KL}}$ increases it from $0.27$ to $0.94$. This difference follows from the return structure. Entropy decrease is signed, so later convergence can offset a temporary rise, making the accumulated return reflect belief convergence. In contrast, KL divergence is always non-negative and thus rewards large belief changes even when they provide no articulation information. For example, particle-filter resampling restores uniform weights, producing a large divergence to which belief entropy is much less sensitive. Joint displacement measures how far the part moves but not whether the motion distinguishes the hypotheses. Overall, $\Delta H$ rewards efficient exploration that gathers useful information during interaction.
\begin{table}[!t]
\centering
\caption{\textbf{Reward ablation on manipulation success rate.}}
\label{tab:info_reward_ablation}
\setlength{\tabcolsep}{2.5pt}
\renewcommand{\arraystretch}{1.2}
\scriptsize
\begin{tabular*}{\columnwidth}{@{\extracolsep{\fill}}lccccc@{}}
\toprule
\multirow{2}{*}{\textbf{Variant}}
& \multicolumn{2}{c}{\textbf{Door}}
& \multicolumn{2}{c}{\textbf{Drawer}}
& \multirow{2}{*}{\textbf{ArticuRiddle}} \\
\cmidrule(lr){2-3}\cmidrule(lr){4-5}
& Train & Val
& Train & Val & \\
\midrule

w/o RL
& 57.6$\pm$1.2
& 40.2$\pm$2.4
& 82.7$\pm$3.0
& 87.7$\pm$1.0
& 48.1$\pm$1.1 \\

$\delta q$
& 63.9$\pm$1.5
& 42.9$\pm$4.2
& 84.4$\pm$1.0
& 90.4$\pm$2.3
& 55.2$\pm$2.8 \\

$D_{\mathrm{KL}}$
& 64.4$\pm$1.1
& 42.9$\pm$3.2
& 87.1$\pm$3.9
& 93.9$\pm$1.3
& 58.6$\pm$4.6 \\

$\Delta H$ (Ours)
& \textbf{68.2}$\boldsymbol{\pm}$\textbf{1.3}
& \textbf{51.3}$\boldsymbol{\pm}$\textbf{1.9}
& \textbf{89.0}$\boldsymbol{\pm}$\textbf{3.0}
& \textbf{95.4}$\boldsymbol{\pm}$\textbf{1.5}
& \textbf{61.7}$\boldsymbol{\pm}$\textbf{1.1} \\

\bottomrule
\end{tabular*}
\end{table}

\subsection{Real-World Evaluation}
For hardware evaluation, we retrain the pipeline in simulation with a 6-DoF YAM arm and deploy it without real-world fine-tuning. The system comprises a YAM arm, parallel-jaw gripper, and RealSense D455 camera, mounted ${\sim}0.65\,\mathrm{m}$ above and in front of the cabinet and extrinsically calibrated using an eye-to-hand ChArUco procedure. World-frame end-effector targets are converted into joint targets using the same damped least-squares IK as in simulation and executed by a safety-constrained controller with displacement clipping and command smoothing. The simulated camera uses the extrinsics while randomizing camera pose, calibration bias, and depth noise. Belief updates take $12\,\mathrm{ms}$ with constant per-step cost; flow is evaluated analytically at every observed point.

SAM~2~\cite{ravi2025sam} segments the target part using the calibrated handle position projected into the first image. Since a prismatic-object mask may include the entire drawer box, a local surface-normal test retains the moving front face and provides its sliding direction, more reliably than the visible centroid, which can drift with the observed region. The mask is lifted through pixel correspondences, cropped to the workspace, and reduced to $4000$ points by farthest-point sampling. For Eq.~\eqref{eq:posterior}, trimmed nearest-neighbor ICP registers the initial masked cloud to each subsequent observation, warm-started from the accumulated transform; referencing the initial frame prevents compounding drift. The resulting rigid motion supplies the part-centroid displacement for revolute hypotheses and the front-face normal for prismatic hypotheses.
\begin{figure}[t]
    \centering
    \includegraphics[width=\linewidth]{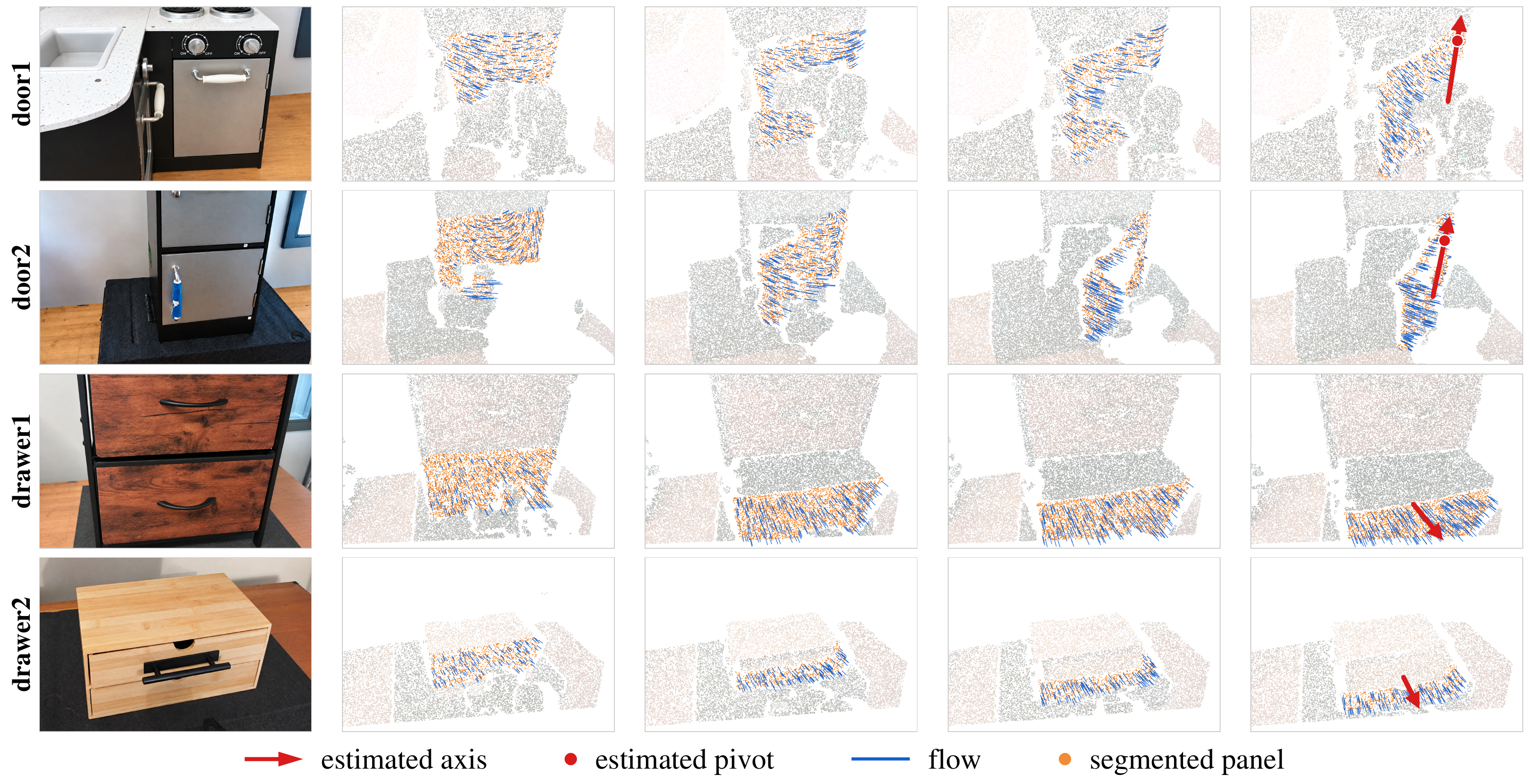}
    \caption{Real-world manipulation examples. Each row shows the
    object followed by point-cloud visualizations at successive
    interaction steps. The rendered flow field shows how observed
    part motion drives the articulation belief to converge toward
    the true joint.}
    \label{fig:realworld}
\end{figure}
We test unseen objects with revolute and prismatic joints varying in size, appearance, handle configuration, and hinge side (Fig.~\ref{fig:realworld}), conducting $10$ trials per object with varied poses and arm reinitialization. From the closed-object point cloud, the panel edge opposite the handle defines the reference revolute hinge, while the horizontal handle-to-base direction defines the prismatic axis. As reported in Table~\ref{tab:realworld} under its captioned criterion, the prior is obtained by running NAP once on a mesh extruded from one closed-state RGB-D frame, without annotation or interaction. Interaction reduces initial axis errors to $1.7$--$3.6^\circ$, recovering axes within a few degrees and matching the simulation trend in Table~\ref{tab:articulation_estimation}. Success reaches $65\%$ for revolute and $85\%$ for prismatic trials. Consistent with simulation, prismatic motion immediately constrains translation, whereas resolving a distant revolute pivot requires a sufficient swept arc. Six failures arise from the gripper not closing firmly on the handle, and four from mask leakage or late-episode self-occlusion during opening. These results demonstrate transfer without real-world training data.
\begin{table}[t]
    \centering
    \caption{Real-world articulation estimation errors and manipulation
    success rates. A trial succeeds when the door opens by more than
    $30^\circ$ or the drawer by more than $20\%$ of its opening range.}
    \label{tab:realworld}

    \small
    \setlength{\tabcolsep}{3pt}
    \renewcommand{\arraystretch}{1}

    \begin{tabular}{@{}lccc@{}}
        \toprule
        Object
        & Axis ($^\circ$) $\downarrow$
        & Pivot (cm) $\downarrow$
        & SR (\%) $\uparrow$ \\
        \midrule
        Door 1   & 2.5 & 2.2 & 60.0 \\
        Door 2   & 3.1 & 2.6 & 70.0 \\
        Drawer 1 & 1.7 & --  & 90.0 \\
        Drawer 2 & 3.6 & --  & 80.0 \\
        \bottomrule
    \end{tabular}
\end{table}
\section{Conclusion}

We introduced a closed-loop framework to represent and reduce uncertainty over
the kinematic structure of articulated objects that cannot be reliably
determined from visual appearance alone. The system includes a particle-based
belief over articulations initialized from a generative prior, a Bayesian
update that refines this belief from the part motion observed during
interaction, a flow field representation that renders the belief into the
policy observation, and an information reward combined with task reward to encourage the policy to gather useful information while manipulating.
Experiments show that our method improve manipulation, outperforming previous work. We then showed that interaction improves the articulation estimate,
increasing joint-type accuracy and reducing axis and pivot errors. The policy also transfers zero-shot to four real-world objects while refining its articulation estimate and flow. Our work leaves several directions for future research. At
present, we consider single-DoF joints covered by
the initial hypothesis set, and the policy relies on a handle to grasp the
part. The performance of our method is also limited when the robot fails
to induce observable motion or when tracking is unreliable. It may be useful to
extend our method to more complex and multi-part mechanisms.

\noindent
\textbf{Acknowledgements}
The authors highly appreciate financial support through the grants NSF FRR 2220868, NSF IIS 2212433, and ONR N00014-22-1-2677.

\bibliographystyle{IEEEtran}
\bibliography{references}

\end{document}